\documentclass[abstract,final]{nldl}

\paperID{40}
\usepackage{mathtools}

\usepackage{pifont}
\usepackage{graphicx}      % Include graphics
\usepackage{amsmath}       % Math environments
\usepackage{amssymb}       % More math symbols
\usepackage{algorithmic}   % Algorithms
\usepackage{algorithm}     % Algorithm float
\usepackage{textcomp}      % Extra symbols
\usepackage{lipsum}

\usepackage{graphicx}

\usepackage{booktabs}

\usepackage{enumitem}

\usepackage{algorithm}
\usepackage{algorithmic}

\usepackage{listings}
\usepackage{hyperref}
\usepackage{url}
\hypersetup{
  pdfusetitle,
  colorlinks,
  linkcolor = BrickRed,
  citecolor = NavyBlue,
  urlcolor  = Magenta!80!black,
}

\usepackage[capitalise]{cleveref}

\usepackage{pifont}

\title{How Many, but How Sure? Class Competition as a Confidence Proxy in Multi-Class Density-Based Counting}
\author[1,2]{Villanelle O'Reilly\thanks{Corresponding Author.}\thanks{This work was supported by the UKRI AI Centre for Doctoral Training in Sustainable Understandable agri-food Systems Transformed by Artificial INtelligence (SUSTAIN), with the grant reference: EP/Y03063X/1.}}
\author[3]{Georgios Leontidis}
\author[1]{Marc Hanheide}
\affil[1]{Lincoln Centre for Autonomous Systems, University of Lincoln, UK}
\affil[2]{Applied Vision and Artificial Intelligence Lab, University of Lincoln, UK}
\affil[3]{UiT The Arctic University of Norway, Norway}
\affil[\ding{41}]{\texttt{\{voreilly, mhanheide\}@lincoln.ac.uk}, \texttt{georgios.leontidis@uit.no}\textsuperscript{\ding{41}}}

\begin{document}
  \maketitle

  \begin{abstract}
    Multi-class density estimators can count objects in crowded scenes, but their class-wise outputs offer no indication of how reliably density mass has been allocated. We introduce a training-free confidence proxy based on local competition between the strongest class channels. The resulting prominence measure requires only the predicted density maps, with no retraining or access to model internals. We test whether residual competition is diagnostic of cross-class leakage by attenuating contested density at inference time. On DSACA predictions for the Hicks et al. dataset, this reduces overall MAE from 0.920 to 0.846, suggesting that class competition contains useful information about count reliability.
  \end{abstract}

  \section{Introduction}

  Crowded environments pose a challenging counting task for humans and computers alike. Such environments arise when surveying flowers in the field for biodiversity monitoring, or when counting people and different vehicle types from overhead, and remote-sensing imagery to inform urban planning. These domains are inherently multi-class, and exhibit sufficiently non-uniform congestion that the assumption of individual discreteness is invalidated by occlusion and proximity between instances, undermining the most capable discrete counting methods, as reported in \cite{gomez2021deep,xu2021dilated,michel2022class,oreilly2026getnuri}. Therefore, estimating a continuous density map can be a more appropriate modelling choice than discretising by counting individual detections or instance masks. However, very little computer vision research has delivered both continuous counting and class discrimination robust to the epistemic uncertainty and modelling problems inherent to cluttered, class-diverse in situ domains.

  Per-class density maps are not mutually exclusive, so several classes may incorrectly claim the same density mass from the same evidence at a given location---this class competition is a direct source of uncertainty in the class-wise counts. Explicit confidence-awareness is absent from multi-class density-based counting, yet it is particularly needed there as a count-class allocation is difficult to act upon without some indication of its reliability. A biodiversity survey or a planning decision needs to know not only how many targets are present, but how confidently count mass has been assigned to classes. A measure of confidence in the class assignment of density-based counts would close this gap.
  
  We hypothesise that regions with both high density activity and strong class competition contain more allocation error than high-activity regions dominated by one class. To test this hypothesis, we introduce the \emph{prominence ratio}, a training-free, spatially resolved confidence proxy computed solely from predicted density maps. We evaluate its diagnostic value by using it to attenuate contested density in frozen multi-class counters and measuring the resulting change in count error.

  \section{Related Work}

  \begin{figure*}[t!]
    \includegraphics[width=\linewidth]{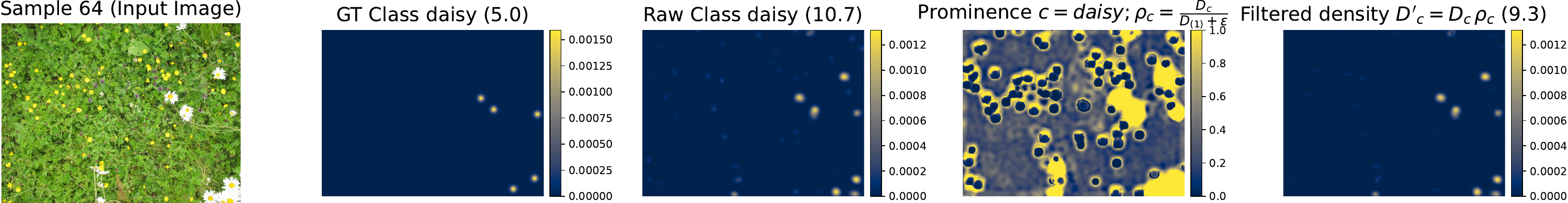}
    \caption{Qualitative sample from the \textcite{hicks2021deep} dataset, using DSACA \cite{xu2021dilated}; weights from GeTNuRi \cite{oreilly2026getnuri}.}\label{fig::qualitative}
  \end{figure*}

    Uncertainty can be modelled directly within a density-based counter. The single-class \textcite{tarling2022deep}, for example, predict an image-level noise variance alongside their single-class density map, optimised through an uncertainty-regularised count loss so that uncertainty in the target counts is learned as part of the regression task itself. Nothing comparable has been attempted for the multi-class case, where a distinctive source of uncertainty is cross-class competition---activations in one class channel contributing count mass that belongs to another. Such channel cross-talk is also a broader quantity than the observation noise Tarling et al.\ model, reflecting a distinctive source of uncertainty in the multi-class setting.

    Rather than explicitly model this uncertainty, multi-class density counters compensate for competition indirectly, through an auxiliary supervised task. \textcite{michel2022class} pair the density estimator with a detection-and-classification pipeline, while DSACA \cite{xu2021dilated} and GeTNuRi \cite{oreilly2026getnuri} attach a non-exclusive, segmentation-like head trained with binary cross-entropy, via their \textit{Category Attention Module} and \textit{Category Focus Module} respectively. In GeTNuRi the auxiliary loss is shown empirically to aid the density optimisation, reducing channel competition over training. Closest to our concern, NWPU-MOC \cite{gao2024nwpu} penalises competition explicitly, regularising the cosine similarity between each pair of flattened class channels. In every case, however, competition is treated as a nuisance to be suppressed at training time, either as a by-product of a borrowed objective or as a whole-image regulariser, rather than located---it still does not ask where class mass is contested, nor does it address whether contested mass is in fact erroneous. No existing method exposes the competition that remains at inference as a measure of confidence in the class assignment of count mass, which is the gap this paper addresses.
    
    \section{Confidence} \vspace{-0.5em}

    A multi-class density counter maps an image $I$ to a non-negative density map $D = \mathcal{F}(I;\theta) \in \mathbb{R}_{\geq 0}^{C\times H \times W}$, one channel per class, with the predicted count for class $c$ given by $\hat{n}_c = \sum_{h,w} D_{c,h,w}$. Nothing constrains the raw mass at a location to belong to a single class, and where several channels are active at once, their counts are in competition for the same evidence. We ask whether the degree of that competition predicts where its class-wise counts are wrong.

    At each location $(h,w)$, let $D_{(1)}$ denote the largest activation across the $C$ channels. We define the \emph{prominence ratio} of each class \begin{equation}\rho_{c,h,w} = \frac{D_{c,h,w}}{D_{(1),h,w} + \epsilon} \in [0, 1],\end{equation} which is $1$ for the locally prominent class and decreases toward $0$ for channels the model has left comparatively unsupported for the evidence at that point. $\rho$ is a property of a class at a location, not a calibrated probability; only a description of how the model has distributed density mass across its channels at a point, computed without labels, training, or access to the model's internals.
    
    To test the hypothesis directly we apply the simplest intervention consistent with it, attenuating each channel by its own prominence \begin{equation}D'_{c,h,w} = D_{c,h,w}\,\rho_{c,h,w},\end{equation} which leaves the locally prominent class untouched, scaling less prominent mass down in proportion to how prominent it is.

    \Cref{fig::qualitative} illustrates this behaviour for the daisy class. Faint daisy activations in the unfiltered prediction largely coincide with stronger buttercup density and are attenuated by prominence filtering. By contrast, isolated daisy peaks retain prominence values near $1$, preserving that prominent predicted daisy mass. The example suggests that low prominence can localise cross-class leakage, although it does not establish that all spatially overlapping activation is erroneous.

    Prominence filtering reduces test MAE for GeTNuRi by 4.5\% on Hicks and by 2.7\% on iSAID. For DSACA on Hicks, it reduces MAE 8.0\%, from 0.920 to 0.846. Improvements across two architectures and two visual domains suggest that residual class competition carries useful information about the reliability of class-wise density allocation. However, filtering deteriorates the test MAE of GeTNuRi on VisDrone by 0.87\%. Low prominence is therefore not synonymous with error---contested density may also represent valid co-occurrence or differences in the relative scaling of class channels. We consequently interpret prominence as a diagnostic confidence proxy rather than a guaranteed correction rule.
    
  \section{Conclusion}

    We introduced the \emph{prominence ratio}, a spatially resolved, training-free proxy for confidence in class-wise density allocation, and tested whether residual competition identifies unreliable density by attenuating low-prominence mass in frozen counters. Filtering reduces MAE for DSACA on Hicks and for GeTNuRi on iSAID and Hicks. Its slight degradation on VisDrone shows, however, that competition is informative but not invariably erroneous---contested mass may reflect either cross-class leakage or valid co-occurrence. Future work should validate prominence directly against spatial allocation error and establish it as a metric capable of distinguishing these cases. As a differentiable quantity, prominence may also support competition-aware training and, ultimately, confidence-aware, prompt-faithful open-vocabulary counting \cite{ciampi2025mindtheprompt}.

  \printbibliography

\end{document}